\documentclass[letterpaper]{article} 
\usepackage{aaai25}  
\usepackage{times}  
\usepackage{helvet}  
\usepackage{courier}  
\usepackage[hyphens]{url}  
\usepackage{graphicx} 
\usepackage{natbib}  
\usepackage{caption} 
\usepackage{algorithm}
\usepackage{algorithmic}

\usepackage{booktabs} 
\usepackage{amsmath}
\usepackage{xcolor}
\usepackage{float}
\usepackage{amssymb}
\usepackage{array}    
\definecolor{orangered}{rgb}{1.0, 0.27, 0.0}

\usepackage{url}

\usepackage{newfloat}
\usepackage{listings}
\DeclareCaptionStyle{ruled}{labelfont=normalfont,labelsep=colon,strut=off} 
\floatstyle{ruled}
\newfloat{listing}{tb}{lst}{}
\floatname{listing}{Listing}
 \nocopyright

\title{Large Language Models for Multimodal Deformable Image Registration}
\author{
    Mingrui Ma\equalcontrib,
    Weijie Wang\equalcontrib,
    Jie Ning,
    Jianfeng He\thanks{Corresponding Authors},
    Nicu Sebe\footnotemark[2],
    Bruno Lepri\footnotemark[2]\\
}
\affiliations{
mamr@kust.edu.cn, weijie.wang@unit.it, njninja@stu.kust.edu.cn,\\ jfenghe@kust.edu.cn, niculae.sebe@unitn.it, lepri@fbk.eu
}

\usepackage{bibentry}

\begin{document}

\maketitle

\begin{abstract}
The challenge of Multimodal Deformable Image Registration (MDIR) lies in the conversion and alignment of features between images of different modalities.
Generative models (GMs) cannot retain the necessary information enough from the source modality to the target one,
while non-GMs struggle to align features across these two modalities. 
In this paper, 
we propose a novel coarse-to-fine MDIR framework, \emph{\textbf{LLM-Morph}}, which is applicable to various pre-trained Large Language Models (LLMs) to solve these concerns by aligning the deep features from different modal medical images.
Specifically, we first utilize a CNN encoder to extract deep visual features from cross-modal image pairs, then we use the \emph{first} adapter to adjust these tokens, and use LoRA in pre-trained LLMs to fine-tune their weights, both aimed at eliminating the domain gap between the pre-trained LLMs and the MDIR task.
Third, for the alignment of tokens, we utilize other \emph{four adapters} to transform the LLM-encoded tokens into multi-scale visual features, generating multi-scale deformation fields and facilitating the coarse-to-fine MDIR task.
Extensive experiments in MR-CT Abdomen and SR-Reg Brain datasets demonstrate the effectiveness of our framework and the potential of pre-trained LLMs for MDIR task. Our code is availabel at: \url{https://github.com/ninjannn/LLM-Morph}.
\end{abstract}

\section{Introduction}

\textbf{M}ultimodal \textbf{D}eformable \textbf{I}mage \textbf{R}egistration (MDIR) aligns analogous anatomical structures in two images from distinct imaging modalities, thereby offering complementary anatomical and functional insights \cite{Multimodal-Registration_Review_2022}. This capability is crucial for various applications, including precise tumor localization \cite{tumorlocate}, organ transplantation assessment \cite{Orgtransplant_2021}, and the segmentation and functional analysis of anatomical structures \cite{anatomicalFunction_2020}.


Traditional MDIR methods \cite{Traditional_2015_ISBI} use iterative optimization algorithms for image alignment but often face time-consuming issues and local optima \cite{TraditionalMethodsIssue}. Recently, deep learning-based methods, including CNNs and Transformers, significantly improve computational efficiency and reduce computing times \cite{Introduction_DeepLearning_2020}, facilitating the MDIR field.
Some MDIR works use
Generative Models (GMs) like SymReg-GAN \cite{SymReg-GAN}, DualStream-GAN \cite{DualStreamGAN}, and TarGAN \cite{TarGAN} to transform images between different modalities using monomodal similarity metrics, avoiding the challenge of multimodal similarity measurement. 
However, when GMs are employed to translate source information to the target, critical source-specific information may be lost during transformation, potentially compromising the quality of the warped image \cite{ContraReg_MICCAI_2022}.
In contrast, non-GMs \cite{VoxelMorph,TransMorph_MIA_2022,MambaMorph_2024} are utilized to predict an explicit deformation field without modal translation.
Yet most non-GMs predict the single-scale deformation field, which cannot handle the large deformation prediction \cite{LapIRN}.
In addition, non-GMs employ information-based similarity metrics, such as mutual information \cite{MI_2019} and MIND \cite{MIND_2012} to learn the voxels correspondence between different modalities.
Although these similarity metrics can be measured independently of modality-specific information, the deep learning-based models still struggle to unify features across different modalities~\cite{Geometry_Q2_2023}.
Consequently, their limited representational capacity, which is more suited to single-modal rather than multimodal data, constrains the performance of the MDIR task~\cite{similarity_metrics_2020}.


Recently, Large Language Models (LLMs) \cite{GPT4_2023,Phi3_2024,LLaMA2_2023} have attracted widespread attention due to their rich corpus knowledge and multi-task capabilities. 
In language-vision tasks, LLMs can understand textual information and associate it with visual data, demonstrating their ability to comprehend and align features across both language and vision modalities.
Recent research \cite{LLMinVision,LLMinVision3} indicates that employing lightweight fine-tuning strategies, such as Adapter \cite{Adapter_2019} and Low-Rank Adaption (LoRA) \cite{Lora_2021}, enhances the versatility of pre-trained LLMs for tasks across various modalities. 
These strategies significantly improve the adaptability of LLMs.
The exceptional cross-modal capabilities of LLMs offer a new viewpoint
to address the challenges faced by GMs and non-GMs in MDIR tasks.

To solve voxel loss caused by GMs during the sampling process and the difficulty that non-GMs face in unifying features of different modalities, 
we propose an coarse-to-fine MDIR method based on pre-trained LLMs, \textbf{\emph{LLM-Morph}}. 
By leveraging pre-trained LLMs as an intermediate modality, we aim to achieve deep feature alignment of two modal images at the same semantic level through pre-trained LLMs. 
However, most current LLMs focus on tasks such as natural language processing or vision-language, which creates a domain gap with MDIR tasks, preventing LLMs from directly performing MDIR.
To address this problem, we construct LLM-Morph using a trainable CNN-based encoder, five adapters, and two blocks with pre-trained LLM layers.
First, the CNN encoder extracts deep features from a pair of images of different modalities. 
Next, outside the LLMs, these features are tokenized and adjusted to the dimensions required by LLMs through the first adapter, addressing the domain difference by projecting these tokens. To further close this domain gap and enhance feature alignment, we fine-tune the pre-trained weights using LoRA within the LLMs.
In the decoding stage, we use four adapters to simultaneously map the LLM-aligned multimodal tokens to the dimensions required for each stage, and then restore these tokens to visual features at each resolution level. 
These visual features are directly utilized to generate deformation fields, thereby achieving multi-scale MDIR from coarse to fine. 
Additionally, we investigate the impact of different layers and various LLMs on the specificity and performance of this task. To the best of our knowledge, this is the first work to employ LLMs for the MDIR task.
Our main contributions are as follows:

\begin{itemize}
\item We develop a novel multi-scale registration framework, LLM-Morph, which utilizes LLMs, to facilitate the alignment of multimodal images' features.
\item We use adapter and LoRA to adjust the features and the pre-trained weights outside and inside the LLMs, respectively, to eliminate the domain gap between the pre-trained LLM and MDIR.
\item We conduct extensive experiments including testing the performance of different pre-trained layers of LLMs in LLM-Morph. In addition, we also test the impact of some pre-trained LLMs on the performance of MDIR.

\end{itemize}

\section{Related Work} \label{related work}
\subsection{Learning-Based Multimodal Registration Methods}

Currently, learning-based MDIR approaches can be roughly divided into two types: \emph{GM-based} and \emph{non-GM-based}. 

For GM-based methods, the QACL framework \cite{QACL_2023_MIA} integrates a quadruple attention generative adversarial network with a monomodal registration network through closed-loop learning. The quadruple attention mechanism effectively addresses the problem of insufficient feature extraction in traditional models. DiCyc \cite{DiCyc_2021} is a cycle-consistent CycleGAN model that tackles domain-specific deformation issues by introducing a global transformation model and enhanced deformable convolution layers. DiffuseMorph \cite{DiffuseMorph_ECCV_2022} utilizes a diffusion-based model to address the topology preservation challenge in continuous deformation and combines it with feature scaling techniques to enhance the flexibility of image registration and reduce computational cost.
However, these methods share a common disadvantage: during the process of translating the source modality (modality of moving image) to the target modality (modality of fixed image), it is inevitable that voxels will lost some key information. The translated moving images lose detailed information, leading to error accumulation in the subsequent registration process and ultimately resulting in inaccurate registration outcomes.

In contrast, non-GM-based methods focus directly on image alignment, thereby avoiding the need for modality translation. For instance, AIR-Net \cite{Weakly_supervisedconvolutional_2018}, is a weakly supervised method based on CNN, which uses high-level correspondence information from anatomical labels to solve the voxel-level spatial correspondence challenge in MDIR. However, the simple combination of conventional convolution and pooling operations cannot align the complex features of multi-modality. ms-RNet \cite{MIND_MI_LOSS_2023} solves the problem of modality difference by combining the loss function of global mutual information and local MIND similarity. However, the limitations of MIND make the final registration result dependent on the initial alignment and the clarity of local features. For complex features, it may blur them and miss structural details. Mamba module \cite{mamba_2023} is a state-space model with linear computational complexity that is highly efficient for high-resolution vision tasks. MambaMorph \cite{MambaMorph_2024} uses the Mamba module to accurately capture fine-grained long-range correspondences and establish voxel-level correspondences between different modalities. However, the inherent 1D nature of the Mamba module may lead to the loss of spatial information when processing 3D medical images \cite{Vision_Mamba_2024}. Although non-GM-based methods can directly predict the cross-modal deformation field, they still face the challenge of converting and aligning heterogeneous features between different modalities.

\subsection{Large Language Models}

\begin{figure*}[ht]
    \centering
    \includegraphics[width=0.90\linewidth]{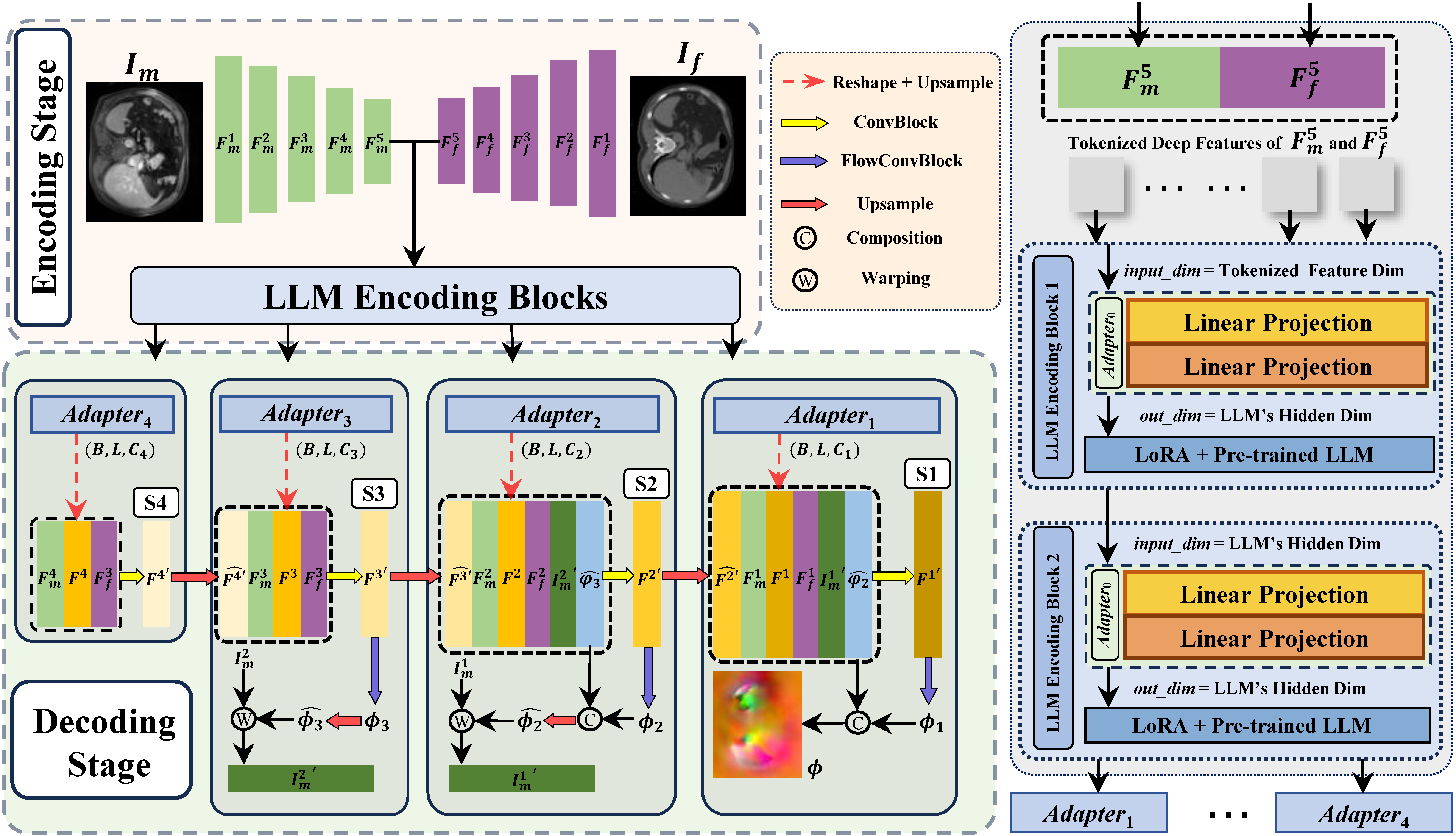}
    \caption{The overall architecture of the proposed LLM-Morph (left) and the details of LLM Encoding Blocks (right).}
    \label{fig:LLM-Morph}
\end{figure*}

GPT-1 \cite{GPT_1} and BERT \cite{Bert_2018} took the lead in expanding the scale of models, marking a new stage in the field of natural language processing and promoting the trend of developing larger-scale models, such as LLaMA 2 \cite{LLaMA2_2023}, LLaMA 3\footnote{\url{https://github.com/meta-llama/llama3}}, Phi3  \cite{Phi3_2024}, and Qwen2  \cite{Qwen2}. LLaMA 2 adopts a decoder-only structure, replacing LayerNorm in the Transformer with RMSNorm, Multi-Head Attention with GQA, and positional encodings with Rotary Embedding. LLaMA 3 uses the same model architecture as LLaMA 2 but utilizes larger-scale pre-training data and changes some parameters. Phi3 uses block sparse attention in the model, introduces sparse patterns on key-value caches, and uses high-quality data for training to achieve performance comparable to larger models with fewer parameters. Compared with Qwen1.5, Qwen2 uses Grouped Query Attention (GQA), reduces the number of decoding layers, attention heads, and Key-Value attention heads, but increases the intermediate size parameter in the MLP. Moreover, these large models not only perform well in natural language processing but also show wide application potential and excellent performance in other fields. M3D-LaMed \cite{LLMInMedical_CVPR_2024} explored for the first time a new path for large language models in 3D medical image analysis and achieved excellent performance on the 3D multimodal medical dataset M3D-Data.

\section{Method}

MDIR can be formulated as an optimization problem based on a similarity measurement between a pair of images in different modalities.
Both images are defined within a three-dimensional spatial domain. The primary optimization objective is to determine an optimal deformation field that maximizes alignment between the two images, and the optimization problem can be formulated as:
\begin{equation}
\underset{\phi}{\mathrm{argmin}} \, E(I_m, I_f, \phi) = E_{\text{sim}}(I_m \circ \phi, I_f) + \lambda R(\phi),
\label{eq:goal}
\end{equation}
where $I_m$ and $I_f$ are the input moving image and fixed image respectively, and $\phi$ represents a deformation field, which describes the magnitude and direction of the spatial pixel transformation from $I_m$ to $I_f$. $\circ$ represents the interpolation operation, and $I_m \circ \phi$ represents the moving image warped via the deformation field $\phi$. We follow Eq. \ref{eq:goal} to train our model in a semi-supervised manner.  
Specifically, we adopt Dice loss for \(E_{\text{sim}}(\cdot, \cdot)\), which is utilized for calculating the similarity between the warped segmentation maps of a moving image $S_m \circ \phi$, and the segmentation maps of a fixed image $S_f$.
\(R(\cdot)\) is the L2 regularization term used to enhance the smoothness of the deformation field \cite{VoxelMorph}; while 
\(\lambda\) is a hyperparameter used to balance the role of similarity loss and regularization term. Thus, the semi-supervised training loss function of this work is $L = E_{sim}(S_m \circ \phi, S_f) + \lambda R(\phi) $.

\subsection{Architecture of LLM-Morph}

The overall architecture of our proposed LLM-Morph is shown in Fig \ref{fig:LLM-Morph}. LLM-Moph consists of a CNN-based feature extractor, two LLM Encoding Blocks (LEBs), and four decoding branches. Unlike conventional non-GMs' architectures, we introduce two LEBs at the bottom of the encoder as an intermediate modality to align the deep features in the two different visual modalities.
Each LEB contains an adapter, $Adapter_0$, consisting of linear projections to adjust them to the input dimension matching a pre-trained LLM, enabling the pre-trained LLM to better and understand the potential correspondence between these features. In the decoding stage, we use four adapter modules to progressively process and transform features in a hierarchical manner, and adjust the deep features aligned by the LLM to the feature dimensions required by each decoding stage.

At the encoding stage, we employ the DualPyramid method \cite{DualPyramid_MIA_2022} to encode and generate five pairs of features at varying scales for moving image $I_m$ and fixed image $I_f$ . These features of $I_m$ and $I_f$ are represented as $F_m^i \in \{F_m^1, F_m^2, F_m^3, F_m^4, F_m^5\}$, and $F_f^i \in \{F_f^1, F_f^2, F_f^3, F_f^4, F_f^5\}$, where $F_m^i$ and $F_f^i$ are the extracted features and downsampled by a factor of $(1/2)^{i-1}$ with \( i \) ranging from 1 to 5, representing the features of full-resolution stage to 1/16-resolution stage.
To make the visual features meet the input requirements of pre-trained LLMs,
the above features are tokenized, meaning the visual features are flattened into tokens of a fixed length of $L = d \times h \times w$, where $d$, $h$, $w$ are the depth, height and width of each patch, respectively.

The decoding stage consists of four sub-stages: S4, S3, S2, and S1, corresponding to the registration process from 1/8-resolution to full-resolution stages. At each stage, an adapter is utilized to transform the aligned deep features into features capable of producing multi-scale deformations. Specifically, the adapter at each stage is responsible for mapping the features output by the last LEB to the feature dimensions required by that stage, thereby gradually recovering features from coarse to fine, allowing precise adjustment of the deformation fields generated in stages S3, S2, and S1.
The output shape of ${Adapter}_i$ is \((B, L, C_j)\), where \(B\) is the batch size, and \(L\) is the length of a token. In this work, \(B\) and \(L\) are fixed. \(C_j\) represents the output dimension of ${Adapter}_i$,
where \(j \in \{4,3,2,1\}\). 
Starting from S4 and continuing to S1, the number of $Adapter_j$ output channels $C_j$ in each stage is 4 times that of the previous stage ($j: 4\to1$).
This setting ensures that at each decoding stage, the adapter can effectively restore the aligned tokens to 3D visual features so that their sizes match \(F_m^{i}\) and \(F_f^{i}\). Finally, through four stages of adapter processing, the corresponding 1/8-resolution \(F^4\), 1/4-resolution \(F^3\), 1/2-resolution \(F^2\) and full-resolution \(F^1\) are obtained for subsequent processing.


First at the S4 stage, \(F_4\), \(F^4_m\), and \(F^4_f\) are concatenated along the channel dimension, and a fused feature \(F'_4\) is produced through a convolution (kernel size 3, stride 1). This fused feature is upsampled to increase the resolution and passed to stage S3, where the initial deformation field \(\phi_3\) is generated by fused feature \(F'_3\) and doubles its resolution to adjust the moving image \(I^2_m\) at half-resolution. The deformed image \(I_m^{2'}\), along with \(\widehat{\phi_3}\) and \(F'_3\), are used to extract the next deformation field. In the S2 stage, the deformation field \(\phi_2\) generated by \(F'_2\) is combined with \(\widehat{\phi_3}\), repeating the warping process from S3. The S1 stage follows the same steps as S2. Finally, the full-resolution \(\phi\) is generated for spatial warping. The Spatial Transformation Network (STN) \cite{STN} is then used to warp the moving image segmentation maps via deformation field $\phi$. This allows the loss function $L$ to be computed, guiding LLM-Morph in learning its weights.

\subsection{LLM Encoding Blocks} \label{LEB}
The key component of LLM-Moprh is the proposed LEBs, as shown in Fig. \ref{fig:LLM-Morph}, each incorporating a pre-trained LLM. 
The proposed LEBs encode the tokenized features and align different modalities by constructing correlations through two consecutive LEBs.
Before being encoded by the pre-trained LLM, the visual features are adjusted to meet the LLM's input requirements. Initially, the image features are converted into a token format known as \emph{Tokenized Deep Features}, which are then processed through a specially designed adapter. These adapters in LEBs are uniformly defined as $Adapter_0$. Each $Adapter_0$ consists of two linear projections equipped with Instance Normalization and LeakyReLU. The purpose of $Adapter_0$ is to map the tokenized features to the number of channels required by LLM and to align the features with the encoding space of an employed LLM. Afterward, each LLM is fine-tuned to maximize the adaptation of the pre-trained weights to the MDIR task. Thus, LEBs that encode tokenized features are constituted.

\subsection{LoRA Fine-Tuning}
 Low-Rank Adaption(LoRA) \cite{Lora_2021} introduces small, and low-rank matrices to the Query ($Q$), Key ($K$), and Value ($V$) components of the self-attention mechanism, enabling model adjustments without a significant increase in parameters. It can be formulated as
\begin{equation}
h = Wx + W_r x = Wx + BAx,
\end{equation}
where $W \in \mathbb{R}^{d \times k}$ is the \emph{Q}, \emph{K}, \emph{V} weight matrix obtained from the pre-trained model, $d$ is the dimension of the output features and $k$ is the dimension of the input features, $x$ is the input token, \(W_r = BA\) is a low-rank update of the weight matrix, where \(B \in \mathbb{R}^{d \times r}\), \(A \in \mathbb{R}^{r \times k}\), And the rank \(r\) is significantly smaller than the minimum value of \(d\) and \(k\). In general, \(W\) is inherited from the pre-trained model, these weights will not receive gradient updates to retain the knowledge learned by the model from large-scale data, and \(W_r = BA\), the two matrices \(B\) and \(A\) will receive gradient updates during the fine-tuning process, enabling the model to adapt to the needs of new tasks without significantly increasing the number of parameters. We use the LoRA fine-tuning method to update and fine-tune the \emph{Q}, \emph{K}, and \emph{V} matrices in the LLM pre-trained layer within each LEB block. Specifically, we add a low-rank weight $W_r$ to achieve fine-tuning of the LLM for the cross-modal registration task. This method is particularly effective for parsing and aligning images from various medical imaging modalities, thereby improving the performance of MDIR.

\section{Experiments}

\begin{table*}[htp!]
\centering
\caption{Comparison results on SR-Reg and Abdomen MR-CT datasets. Higher Dice (\%) results indicate higher registration accuracy. $|J_\phi| \leq 0$ (\%) indicates the percentage of folding voxels in a deformation. Lower HD95 means more accurate edge matching of segmentation maps. The ``Initial'' is the initial alignment results without registration.}
\label{tab:comparison}
\scalebox{.9}{\begin{tabular}{lccccccc}
\toprule
& \multicolumn{3}{c}{Abdomen MR-CT} & \multicolumn{3}{c}{SR-Reg} \\
\cmidrule(lr){2-4} \cmidrule(lr){5-7}
Methods & \multicolumn{1}{p{2cm}}{Dice (\%) $\uparrow$}  & \multicolumn{1}{p{2.3cm}}{$|J_{\phi}| \leq 0 (\%) $ $\downarrow$} & \multicolumn{1}{p{2cm}}{HD95 (mm) $\downarrow$} & \multicolumn{1}{p{2cm}}{Dice (\%) $\uparrow$} & \multicolumn{1}{p{2.3cm}}{$|J_{\phi}| \leq 0 (\%) $ $\downarrow$} & \multicolumn{1}{p{2cm}}{HD95 (mm) $\downarrow$} \\
\midrule
Initial      & $25.50 (15.05)$ & -- & $32.79 (13.50)$           & $62.42(3.29)$ & -- & $3.73(0.41)$           \\
VoxelMorph  & $67.14 (16.37)$  & $2.54 (0.37)$ & $12.93(6.43)$ & $74.65(2.33)$ & $\mathbf{0.31(0.03)}$ & $2.78(0.34)$ \\
TransMorph   & $70.73(11.51)$  & $1.36(0.31)$ & $17.77(10.44)$ & $81.14(2.49)$ & $0.69(0.04)$ & $2.13(0.27)$ \\
MambaMorph   & $78.04(13.76)$  & $1.26(0.36)$ & $13.37(10.34)$ & $83.01(1.76)$ & $0.83(0.03)$ & $1.89(0.23)$ \\
\textbf{LLM-Morph}     & $\mathbf{80.11(14.46)}$  & $\mathbf{1.11(0.19)}$ & $\mathbf{9.89(7.33)}$ & $\mathbf{83.39(2.15)}$ & $0.57(0.04)$ & $\mathbf{1.87(0.28)}$ \\
\bottomrule
\end{tabular}}
\end{table*}

\vspace{-3mm}

\begin{figure*}[ht!]
    \centering
   \includegraphics[width=.95\linewidth]{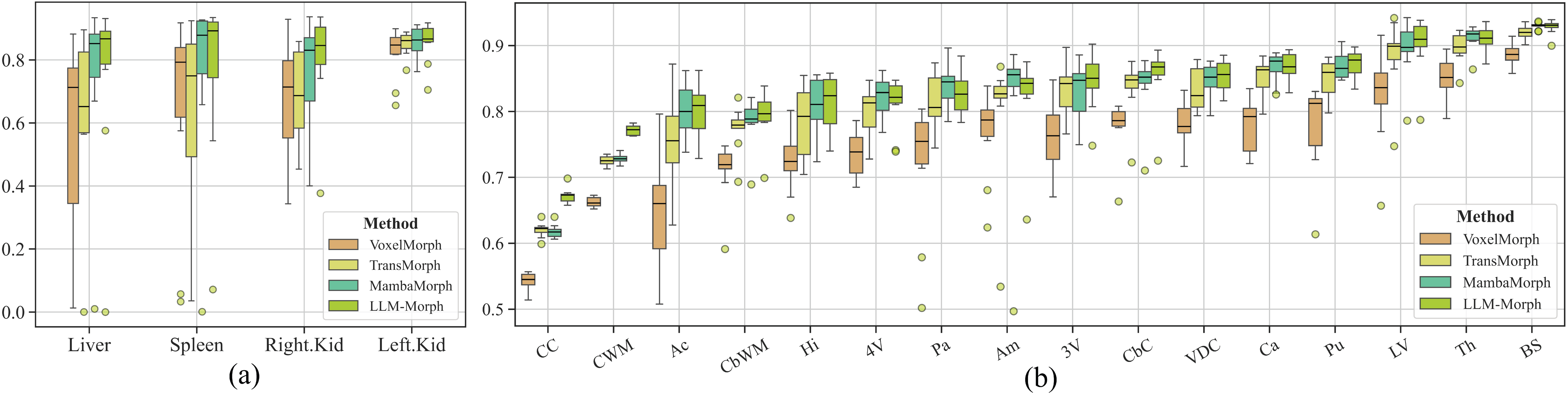}
    \caption{Box plots of the alignment results of all baseline methods on (a) Abdomen MR-CT and (b) SR-Reg datasets. The horizontal axis represents different anatomical structures, and the vertical axis represents the Dice values, respectively.}
    \label{fig:Boxplot}
\end{figure*}

\subsection{Data Preparation}
We evaluate baseline methods and the proposed LLM-Moprh on two public datasets:

\textbf{(1) Abdomen MR-CT dataset} from the Learn2Reg \cite{Learn2Reg_2022_MIA}, comprises three sub-datasets: TCIA, BCV, and CHAOS. It contains a total of 46 MR images and 55 CT images.
All images have corresponding labels and are resampled to $192 \times 160 \times 192$. Following the preprocessing steps in \cite{RelatedWork_CrossAttention_2022_MIA}, we clipped CT images to HU values from -180 to 400 and MR images to the 0 to 95 percentile intensity range. We randomly select 10 images from the MR and CT images to form a fixed 10 pairs of test sets. The remaining 36 MR and 45 CT images are combined in sequence to train the model, ensuring that there are no duplications. Therefore, 1620 pairs for training and 10 pairs for testing are generated.

\textbf{(2) SR-Reg dataset} derived from the preprocessing of the SynthRAD 2023 dataset \cite{SynthRAD_2023}, includes 180 well-affined, skull-stripped, and intensity-normalized brain MR-CT pairs with segmentation maps. 
The main preprocessing steps are brain segmentation with SynthSeg \cite{SynthSeg_MIA_2023}, and skull stripping with SynthStrip \cite{SynthStrip_MIA_2022}.
Following the MamaMorph \cite{MambaMorph_2024} training strategy, we use 150, 10, and 20 pairs for training, validation, and testing, respectively. 
All volumes have a shape of $192 \times 208 \times 176$ voxels with a resolution of $1 \times 1 \times 1 \ mm^{3}$.

\subsection{Implementing Details}
The proposed LLM-Morph is implemented using the PyTorch framework on the NVIDIA GeForce RTX 3090 GPU, utilizing Automatic Mixed Precision (AMP) to increase training speed and reduce memory usage. The output dimension \(C_4\) of $Adapter_4$ is set to 256, and \(r\) is set to 64 for LoRA fine-tuning. We use the Adam optimizer to train LLM-Morph with a learning rate of 0.0001. The batch size $B$ is set to 1 and the hyperparameter $\lambda$ of regularization term is set to 0.1. The equipped LLM layers in LEBs are 15th/16th layers of pre-trained LLaMA-3-8B, with the fixed number of dimensions of input and output are both 4096.

For the MR-CT Abdomen dataset, we utilize half-resolution training strategy to train both baseline methods and LLM-Morph for 50 epochs. Once the corresponding half-resolution deformation field is obtained, it is upsampled to full resolution to warp the segmentation maps and to calculate the loss function, and all methods are trained for 50 epochs. It should be noted that the downsampled image resolution changes from the original $192 \times 160 \times 192$ to a half-resolution of $96 \times 80 \times 96$. Therefore, at the bottom of LLM-Morph, i.e., at the 1/16 resolution stage, the corresponding patch is set to $d=6, h=5, w=6$, resulting in a fixed token length \(L = d \times h \times w = 180\).

For the SR-Reg dataset, both the baseline models and LLM-Morph are trained for 300 epochs using the full-resolution volumes with the shape of $192 \times 208 \times 176$. The corresponding patch size is set to $d=11, h=13, w=12$, resulting in a fixed token length of $L=d\times h \times w = 1716$.
 
\subsection{Evaluation Metrics and Baseline Methods}%

We calculate the Dice and 95\% Hausdorff Distance (HD95) metrics using the warped segmentation maps of the moving image and the original segmentation maps of the fixed image.
Dice and HD95 are both metrics utilized to assess similarity: Dice measures the overlap of corresponding labels in two images, while HD95 evaluates the boundary distance between them.
The percentage of non-positive Jacobian determinant $|J_{\phi}| \leq 0$ to evaluate the foldings during deformation.

To demonstrate the outperformance of LLM-Morph, we compared it with three representative deep learning-based methods: the CNN-based VoxelMorph \cite{VoxelMorph}, the SwinTransformer-based TransMorph \cite{TransMorph_MIA_2022}, and the recently popular Mamba-based \cite{mamba_2023} MambaMorph \cite{MambaMorph_2024}. All methods use consistent hyperparameter $\lambda$, semi-supervised training loss function, and the same data partitioning.

\subsection{Experimental Results}
Table \ref{tab:comparison} shows the quantitative comparison results on the Abdomen MR-CT and SR-Reg datasets. On the Abdomen MR-CT dataset, LLM-Morph significantly outperforms other baseline methods across both metrics. Specifically, LLM-Morph achieves a Dice score that is 2.07\% higher than the second-ranked MambaMorph, 9.38\% higher than the third-ranked TransMorph, and 12.97\% higher than the last-ranked VoxelMorph. 
For the SR-Reg dataset, our method also achieves the best results in both similarity metrics. On the Dice metrics, the ranking of the baseline methods is still consistent with the results on the abdomen dataset, and our method is 0.38\% higher than MambaMoprh, 2.25\% higher than TransMorph, and 8.74\% higher than VoxelMorph. 

In terms of the $|J_{\phi}| \leq 0 (\%) $ metric, our method achieves the best results on both datasets, with the exception of the VoxelMorph results on the SR-Reg dataset. 
We believe that, as can be seen from the Initial of the two datasets in Table \ref{tab:comparison}, the Dice metrics of Abdomen MR-CT are much lower than that of SR-Reg, indicating that the anatomical structure involved in abdomen MR-CT is more complex and the positions of abdominal organs vary greatly. LLM-Morph has different adapters in the decoding stage to restore the aligned features to generate a multi-scale deformation field, which performs well in processing images of organ displacement, deformation, or different density contrasts. The challenges of SR-Reg datasets with relatively simple or uniform displacements are more focused on accuracy and consistency rather than complexity, which may make the performance differences between different algorithms less significant.

Fig. \ref{fig:Boxplot} shows the slice results of each segmentation map on abdomen and brain datasets. Fig. \ref{fig:Boxplot}(a) indicates that LLM-Morph outperforms those of other methods across all anatomical structures on the abdomen dataset. On the brain dataset, as shown in Fig. \ref{fig:Boxplot}(b), LLM-Morph also outperforms other methods in most anatomical structures.
Due to space limitations, the abbreviations of each anatomical structure are provided in the supplementary materials.

\begin{figure}[ht!]
    \centering
    \includegraphics[width=.95\linewidth]{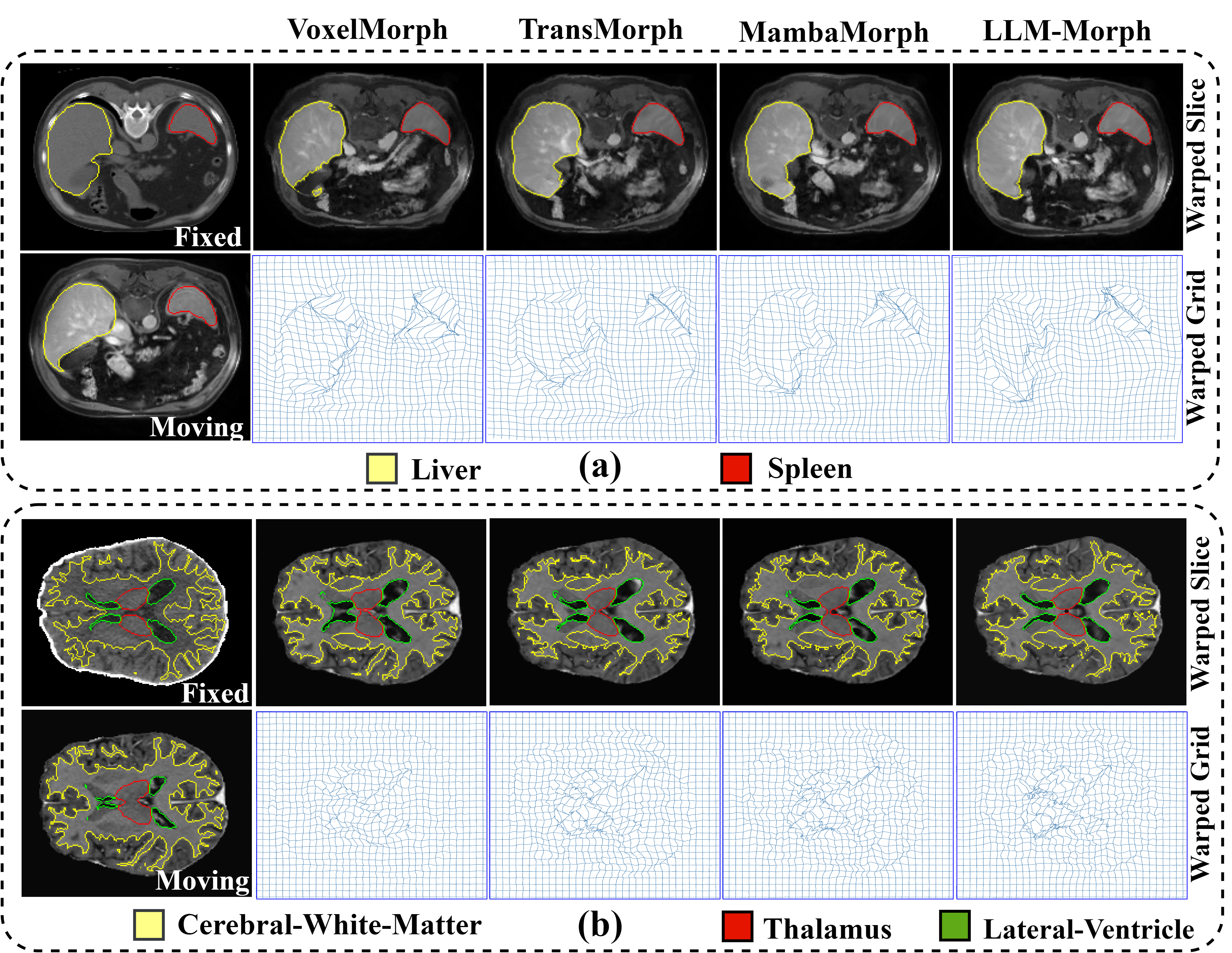}
    \caption{Slice visualization on two datasets. The boundaries in each color represent the edges of different segmentation maps. The warped grid is utilized to observe the deformation field of the current slice, while the warped slice represents the corresponding slice from the warped moving image.}
    \label{fig:Visual_Result}
\end{figure}

The qualitative results are shown in Fig. \ref{fig:Visual_Result}. We can observe in this slice that the warped liver in the moving image is most similar to the one in the fixed image. Fig. \ref{fig:Boxplot}(b) demonstrates that LLM-Morph outperforms other methods on most anatomical structures, especially on CC and CWM labels with lower Dice values, the CC anatomy is shown in the yellow label in Fig. \ref{fig:Visual_Result}(b), which indirectly verifies our effectiveness in dealing with anatomical structures with large displacements while also being able to bridge the distant correlated voxels in different modalities.

\section{Ablation Study} \label{Sec: ab}
We conduct ablation experiments on LLM-Morph to verify the effectiveness of various proposed modules on abdomen dataset for ablations. In Ablation 1, we remove LEBs and the four adapters in the decoding stage. In Ablation 2, we replace the LLaMA 3 blocks in LEBs with the trainable standard ViT \cite{ViT} modules. In Ablation 3, only one LEB is employed. Ablation 4 is LLM-Morph without LoRA fine-tuning.

\begin{table}[ht]\small
\centering
\caption{Ablation results of LLM-Morph. Ablation 1: LLM-Morph has no LEBs and adapters except the inner $Adapter_0$. Ablation 2: LLaMA 3 blocks is replaced with (r/w) the standard ViTs. Ablation 3: LLM-Morph with one LEB. Ablation 4: LLM-Morph without LoRA fine-tuning. }
\label{tab:ab_comparison}
\begin{tabular}{>{\centering\arraybackslash}p{0.8cm}p{3cm}p{0.7cm}>{\centering\arraybackslash}p{1.2cm}p{0.7cm}} 
\toprule
Ablation & Methods & Dice & $|J_{\phi}| \leq 0$ & HD95 \\
\midrule
- & Initial       & 25.50  & -    & 32.79 \\
1 & w/o LEBs and Adapters   & 74.64 & 1.34 & 10.43  \\
2 & LEBs r/w ViTs   & 75.23 & 1.24 & 13.86 \\
3 & w/ one LEB  & 76.64 & \textbf{1.07} & 11.56 \\
4 & w/o LoRA          & 79.44  & 1.23 & 10.19  \\
- & LLM-Morph         & $\textbf{80.11}$ & 1.11 & $\mathbf{9.89}$ \\
\bottomrule
\end{tabular}
\end{table}

Ablation results are shown in Table \ref{tab:ab_comparison}. These four groups of experiments all exhibit different degrees of decline, with Ablation 1 showing the largest decline. 
After removing LEBs and adapters, the Dice value drop to 74.64\%, a 5.47\% decrease from LLM-Morph, and HD95 increase from 9.89 to 10.43, proving the effectiveness of LEBs and adapters in our framework. 
After replacing LLaMA 3 blocks in LEBs with ViT blocks in Ablation 2, the Dice metric drop to 75.23\%, and HD95 reach the worst 13.86, demonstrating that the weights of the pre-trained LLaMA 3 has a positive effect on registration performance.
When the number of LEBs reduced from two to one in Ablation 3, the Dice value also declined to varying degrees, and HD95 increased to a certain extent.
Finally, removing the LoRA fine-tuning in Ablation 4 results in a slight decline in the Dice value, which verifies the fine-tuning effect of LoRA. Although LLM-Morph performs slightly higher than Ablation 3 in $|J_{\phi}| \leq 0$, the difference is not large and remains at the same level.

\subsection{Ablation Results of Different Pre-trained Layers}



It is worth noting that LLMs consist of multiple Transformer layers. Specifically, LLaMA 3, employed in our LLM-Morph, consists of 32 layers of improved Transformer. Each LEB of the proposed LLM-Morph selectively loads only one layer from these 32 layers. To verify whether loading different pre-trained layers affects the experimental results and to identify which layers are optimal for MDIR, we perform extensive validations.

\begin{figure}[ht!]
    \centering
    \includegraphics[width=.9\linewidth]{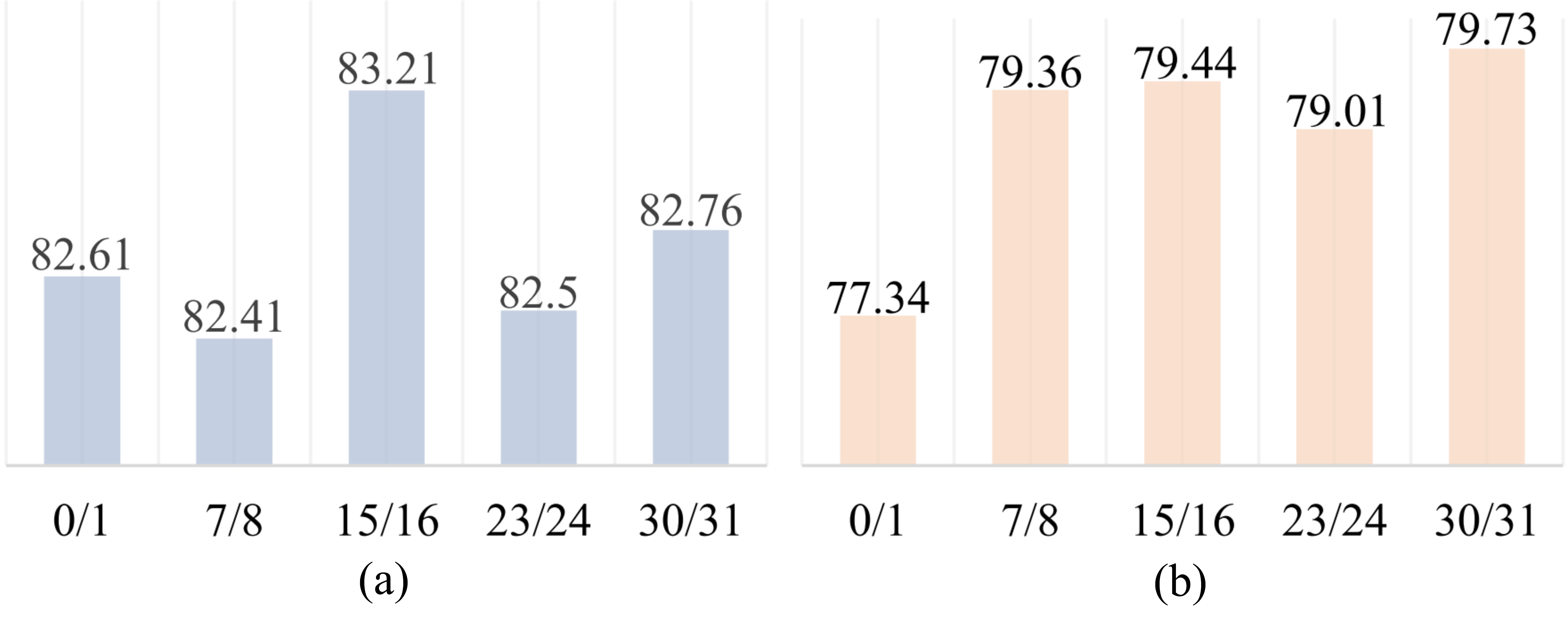}
    \caption{Histogram results of different layers equipped in LLM-Moprh. The horizontal axis indicates the layer ordinal, and the vertical axis represents the average Dice value. (a) Results on abdomen dataset. (b) Results on brain dataset.}
    \label{fig:zhifangtu}
\end{figure}

We conduct five experiments loading weights from the 0th/1st, 7th/8th, 15th/16th, 23rd/24th, and 30th/31st layers of LLaMA 3 into the LEBs without LoRA fine-tuning. These experiments are performed on both datasets, and their Dice metrics were measured, as shown in the histogram in Fig. \ref{fig:zhifangtu}. The horizontal axis represents the ordinal number of the layers loaded into LEBs, while the vertical axis is the average Dice value.

Fig. \ref{fig:zhifangtu}(a) shows the experimental results of the abdominal MR-CT dataset. LLM-Morph loaded with weights of 30th/31st layers obtains the highest Dice value of 79.73\%, followed by weights of 15th/16th layers. Fig. \ref{fig:zhifangtu}(b) shows the experimental results of the brain dataset. LLM-Morph loaded with weights of 15th/16th layers achieve the best Dice metric of 83.21\%, followed by weights of 30th/31st layers. In general, LLM-Morph loaded with weights of 15th/16th and 30th/31st layers weights achieved the top two results on both datasets, while LLM-Morph loaded with weights of 0th/1st layers performed poorly on both datasets. Based on the above results, we speculate that in the training process of LLMs based on Transformers, each layer usually assumes different functions and learning responsibilities, and the first few layers (such as 0th and 1st layers) are usually responsible for learning more general and basic features. For MDIR, these basic features may not be sufficient to capture sufficiently complex information for alignment. The middle and later layers are usually responsible for learning more advanced and high-dimensional features. The middle to deeper layers may be better at understanding complex structures and spatial relationships, which are not easily captured by the earlier layers. These experiments show that Transformers' weights at different layers have a significant impact on the task of MDIR, especially the weights of the middle or the deeper layers may be the best choice.

\subsection{Results of Loading Different LLMs}
The designed LEB is sufficiently flexible to accommodate any LLM. Therefore, we further select other LLMs to integrate into LEBs to explore the impact of different LLMs on MDIR performance.
Specifically, we integrate two middle layers of other pre-trained LLMs into LEBs by adjusting the output dimension of $Adapter_0$ to match the requirements of each specific LLM. 
In this experiment, we remove LoRA fine-tuning to focus solely on the impact of different LLMs in LLM-Morph. The four selected LLMs are: LLaMA 2\footnote{\url{https://huggingface.co/meta-llama/Llama-2-7b-chat-hf}}, Phi3Vision\footnote{\url{https://huggingface.co/microsoft/Phi-3-vision-128k-instruct}}, Qwen2\footnote{\url{https://huggingface.co/Qwen/Qwen2-7B}}, and LLaMA-3-Chinese\footnote{\url{https://huggingface.co/FlagAlpha/Llama3-Chinese-8B-Instruct}}.
LLaMA-3-Chinese is a pre-trained model obtained by retraining with a Chinese database following the release of LLaMA 3. The relevant details of these LLMs have been mentioned in the Related Work section. The hidden dimensions of these four LLMs are 4096, 3072, 3584, and 4096, respectively. We verify their performance on the abdomen dataset, and all these pre-trained LLMs weights are obtained from their open-source websites.

\begin{table}[htbp]\small
\centering
\caption{Comparison results of loading different LLMs in LEBs on Abdomen MR-CT dataset.}
\label{tab:methods_comparison}
\begin{tabular}{@{}lccc@{}}
\toprule
 Methods & \multicolumn{1}{c}{Dice}  & \multicolumn{1}{c}{$|J\phi| \leq 0$} & \multicolumn{1}{c}{HD95} \\
\midrule
Initial        & 25.50  & -    & 32.79 \\
LLaMA 2         & 78.76 & 1.05 & $\mathbf{8.53}$  \\
Phi3Vision    & $\mathbf{80.05}$ & 1.16 & 12.03 \\
LLaMA-3-Chinese  & 79.32 & 1.14 & 10.23 \\
Qwen2          & 76.60  & $\mathbf{0.95}$ & 11.40  \\
LLaMA 3         & 79.44 & 1.23 & 10.19 \\
\bottomrule
\end{tabular}
\end{table}


The results of integrating different pre-trained LLMs in LLM-Morph are shown in Table \ref{tab:methods_comparison}. Qwen2 performs best on the $|J_{\phi}| \leq 0$ measurement, but this value is at a similar level across all methods. Our main focus is on the similarity metrics. In terms of Dice and HD95, the performance of each LLM varies significantly. Except for LLaMA 3, the results of the other LLMs on these two metrics are quite unbalanced. Specifically, Phi3Vision achieves the highest Dice score but the lowest HD95 value, while LLaMA 2 performs best in HD95 but ranks second to last in Dice. Among these methods, only LLaMA 3 achieves relatively balanced outcomes across both metrics.
\section{Conclusion}

In this paper, we propose LLM-Morph, the first coarse-to-fine MDIR approach based on pre-trained LLMs, designed to address the challenge of aligning features across different modalities in non-GMs. Our framework elimates the modal gap between LLMs and MDIR, effectively utilizes LLMs to align multimodal image features through the proposed LEBs. Additionally, adapters at different stages recover these aligned features to generate deformation fields at multiple scales, completing the coarse-to-fine MDIR process. Experimental results on two public datasets demonstrate that LLM-Morph outperforms baseline methods, showcasing the effectiveness of LLMs in MDIR. Furthermore, we explore the impact of loading different pre-trained layers in LEBs on MDIR tasks, providing guidance for similar studies in selecting appropriate pre-trained layers. We also compare the performance of different LLMs in MDIR, providing a basis for LLM selection and architecture design in future research.

\bibliography{aaai25}

\end{document}